\documentclass[letterpaper, 10 pt, hidelinks, conference]{ieeeconf}
\pdfoutput=1
\IEEEoverridecommandlockouts
\usepackage{cite}
\usepackage{amsmath,amssymb,amsfonts}
\usepackage{algorithmic}
\usepackage{graphicx}
\usepackage{textcomp}
\usepackage{hyperref}
\usepackage{comment}
\usepackage{xcolor}
\usepackage{CJKutf8}
\let\labelindent\relax
\usepackage{enumitem}

\usepackage{microtype}

\def\BibTeX{{\rm B\kern-.05em{\sc i\kern-.025em b}\kern-.08em
    T\kern-.1667em\lower.7ex\hbox{E}\kern-.125emX}}

\title{\LARGE \bf iTrash: Incentivized Token Rewards for Automated Sorting and Handling}
\author{Pablo Ortega$^{1}$, Eduardo Castell{\'{o}} Ferrer$^{1,2}$\\
\small{$^1$CyPhy Life, Robotics \& AI Lab, School of Science \& Technology, IE University, Spain}\\
\small{$^2$MIT Connection Science, Massachusetts Institute of Technology, Cambridge, USA}}

\begin{document}
\maketitle
\urlstyle{same}
\thispagestyle{empty}
\pagestyle{empty}
{\let\thefootnote\relax\footnote{{Corresponding author: Pablo Ortega, portega.ieu2020@student.ie.edu}}}

\vspace{-0.35cm}
\begin{abstract}
As robotic systems (RS) become more autonomous, they are becoming increasingly used in small spaces and offices to automate tasks such as cleaning, infrastructure maintenance, or resource management. In this paper, we propose iTrash, an intelligent trashcan that aims to improve recycling rates in small office spaces. For that, we ran a 5 day experiment and found that iTrash can produce an efficiency increase of more than 30\% compared to traditional trashcans. The findings derived from this work, point to the fact that using iTrash not only increase recyclying rates, but also provides valuable data such as users behaviour or bin usage patterns, which cannot be taken from a normal trashcan. This information can be used to predict and optimize some tasks in these spaces. Finally, we explored the potential of using blockchain technology to create economic incentives for recycling, following a Save-as-you-Throw (SAYT) model. 
\end{abstract}

\section{Introduction}
\label{sec:intro}
An essential part of reducing greenhouse gas emissions and mitigating escalating effects of climate change~\cite{Kemp2022} is to reduce over-consumption~\cite{WJRipple2023}. The recently popular {\it circular economy} strategy proposes to reduce over-consumption by optimizing material flows, including by recycling waste~\cite{di2015robust}. However, only 6\% of the climate policies reported by European countries\footnote{European countries: the EU-27, Iceland, Norway and Switzerland.} include some form of circular economy policies and measures, according to a 2024 European Environment Agency (EEA) report~\cite{eea2024}. 

Also according to the EEA, 55\% of municipal waste and 65\% of packaging waste should be prepared for re-use or recycling by 2025, but at least 19 of the EU-27 states will struggle to meet that target~\cite{eea2023}. It is essential to build new technologies to make recycling easier, more consistent, and more accurate.

Increasing the efficiency of waste management remains an important open challenge~\cite{blockchain_waste_survey}. Digital technologies can be found in large-scale waste management processes, for example using an IoT-based solution to track the amount of waste collected in urban containers~\cite{GABRIEL2023e00389}, but these technologies do not directly provide solutions for small-scale use~\cite{SALVIA2021129200}, such as in small office and domestic environments. Furthermore, much of the waste deposited in large municipal containers is sorted in private indoor environments beforehand. As waste generation continues to increase in small office and domestic environments~\cite{smart_homes}, it is essential to increase the efficiency of waste sorting~\cite{LU202229} and recycling~\cite{kiyokawa2022challenges} in small office and domestic environments.

{\it Smart spaces}~\cite{efficiency_smart_homes} are indoor environments (e.g., apartments, offices) that use electronic devices and systems to automate and control functions such as lighting~\cite{smart_plug} or heating~\cite{smart_thermostat}, often through a centralized platform accessible remotely via smartphone application. Smart spaces can potentially improve recycling rates~\cite{hui2023greening}, for example with smart trashcans for better sorting. Some smart trashcans have been developed, including a system that checks the bin's level and alerts the user when the bin is almost full~\cite{smarttrashcan}, but to the best of our knowledge, there are none that improve the rate of correct recycling sorting. This paper proposes a smart trashcan that not only detects waste types to support sorting, but offers automated incentives~\cite{incentive_adoption} for users to adopt the technology and use it correctly.

\subsection{Blockchain for incentivized behavior}
In the last decade, the cryptographic technology ``the blockchain'' has provided new ways to organize collectives without the need of a centralized authority~\cite{Castell_Ferrer_2018,ferrer2018robochain,Ferrer2023ICRA,daos} and to create incentive mechanisms within those collectives~\cite{incentive_blockhain_dao}. However, blockchain technology faces significant challenges in latency, size, throughput, and bandwidth, particularly in applications where quick and reliable data sharing is essential~\cite{Castell_Ferrer_2018}. 

New blockchain protocols such as Ripple (XRP)~\cite{schwartz2014ripple} are designed to overcome these problems, with fast transaction speeds (e.g., less than 5 seconds), low transaction fees, and a highly scalable and stable network~\cite{ripple_advantage}. The Ripple network is well-suited for IoT devices transactions~\cite{ripple_1} and is therefore used in this paper. 

Blockchain-based incentive mechanisms have been proposed in many application domains, for example, using token-based reward systems to promote fitness~\cite{zishan_virani_2022_6990368} or incentivize urban cycling~\cite{cycling_blockchain}. In~\cite{kt}, low-carbon mobility options are rewarded with tokens that can be exchanged for free visits to different attractions in a city, while in~\cite{parentalcontrol}, children are rewarded for doing their homework with tokens that can be spent on TV viewing time~\cite{parentalcontrol}.

In recycling and waste management, blockchain-based technology has been proposed for tracing and tracking waste in smart cities~\cite{tracing}, reliable channelization of waste~\cite{channelization}, protection of waste management documentation~\cite{protection}, and rewarding individuals for garbage removal~\cite{blockchain_waste_managment_system_1}. In~\cite{Esmaeilian2018}, it is proposed that future cities should incentivize recycling by securely and transparently transferring cryptocurrency tokens to waste collectors and participants. In this paper, we present a proof-of-concept to progress towards this aim. 

\section{iTrash}
\label{sec:itrash}
We present iTrash (Incentivized Token Rewards for Automated Sorting and Handling), a robotic system to incentivize correct recycling in domestic and small office environments, using a blockchain-based Waste-to-Reward system~\cite{smarttrashcan2} following a save-as-you-throw model~\cite{incentive_adoption}. We built and deployed a proof-of-concept iTrash that rewards the user via the XRP network and compared it with a normal (control) trashcan in a 5-day real-world experiment in a university common area. 

\begin{figure}[h!]
\centering
\includegraphics[width=\columnwidth]{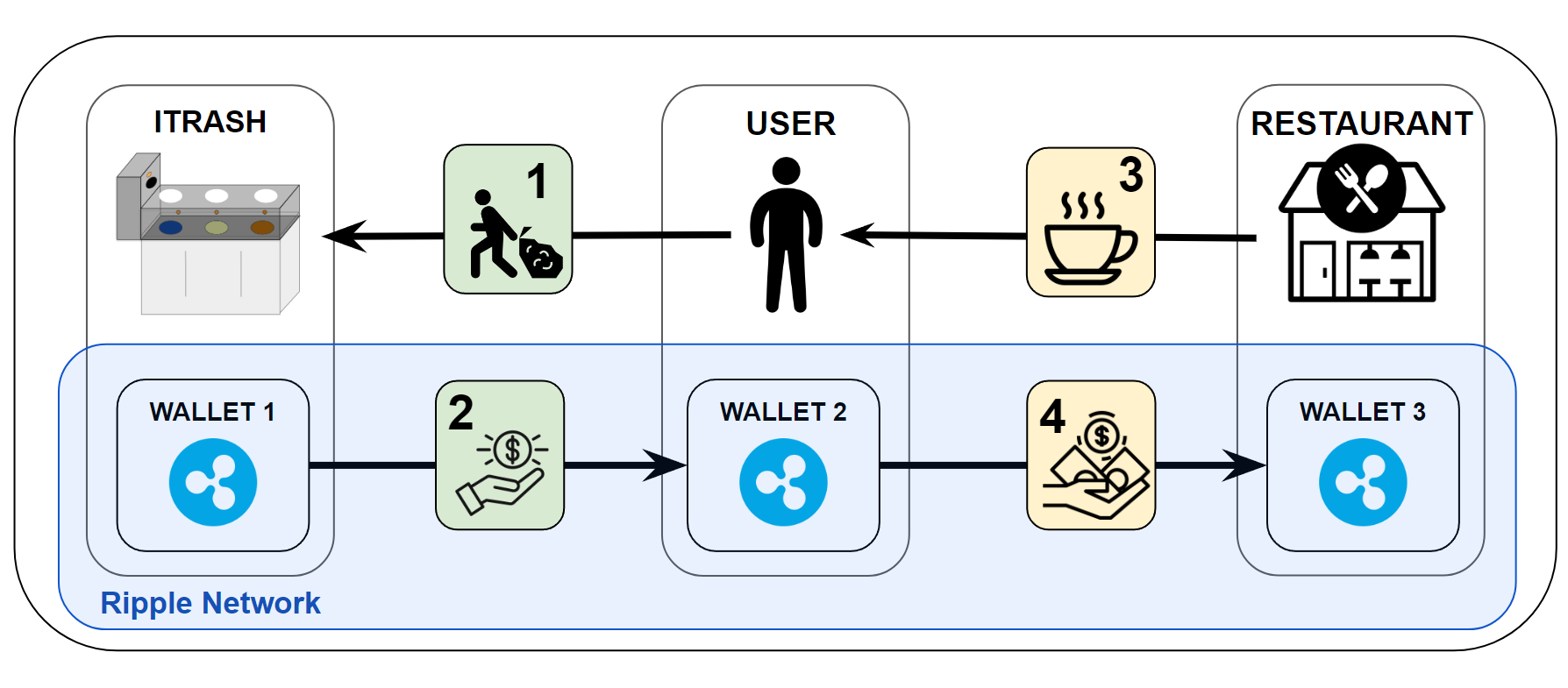}
\caption{{\bf iTrash concept:} Diagram showing the incentive mechanism of iTrash.  When users recycle properly (1), they earn a reward from iTrash (2). Then, they are able to purchase products or services (3) with the rewards (4) received with iTrash.  } 
\label{fig:blockchain}
\end{figure}

Fig.~\ref{fig:blockchain} presents a diagram overview of the iTrash concept, showing how a user would get rewarded. The first step is the recycling step (1), where the user follows all the steps described in section~\ref{ssec:system_overview}. If the user recycles effectively, they will receive a reward in the form of cryptocurrency. This reward will be transferred (2) from the iTrash wallet (Wallet 1) to the user’s wallet (Wallet 2). Once the user has funds in their wallet, they can use them to pay for services or products. For example, consider a restaurant that has its own wallet and accepts XRP as a form of payment. The user can purchase a service (3) and pay for it in XRP (4), transferring the funds to the restaurant’s wallet (Wallet 3). Since the network operates on blockchain technology, every transaction will be transparent and auditable. 

\subsection{System Overview}
\label{ssec:system_overview}
\begin{figure}[b]
\centering
\includegraphics[width=0.9\columnwidth]{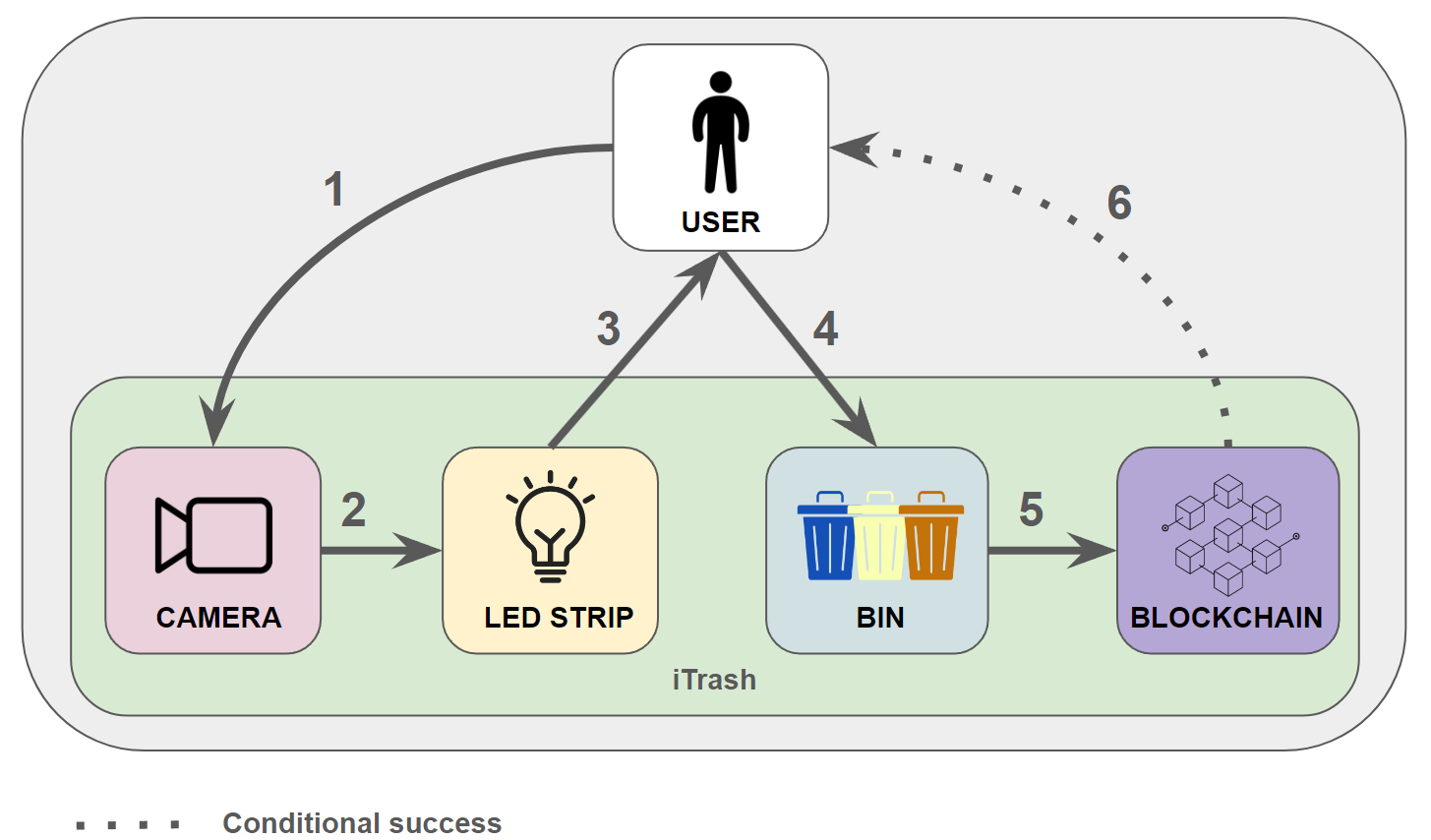}
\caption{High-level vision of system's workflow. 1) User shows a disposal item to a camera within the iTrash system. 2) The camera detects the waste and determines the color of the bin in which to dispose the item. 3) Light displayed from an LED strip with the correct bin color is emitted to inform the user. 4) User throws the item to the corresponding bin. 5) Feedback about this action is sent to a blockchain network. 6) In case the recycle process was successful, the user receives a reward in the shape of a crypto token (i.e., XRP).} 
\label{fig:overview}
\end{figure}

Fig.~\ref{fig:overview} shows the system overview, with a workflow divided in six steps. First, {\bf (1)} the user places a disposable item where it can be detected by a proximity sensor. Once detected, the camera captures an image of the object. Then, {\bf (2)} the image is processed by an image classification model, which outputs the color of the bin that best matches the type of waste the user has shown to the camera. {\bf (3)} An LED strip then illuminates in the color matching the appropriate bin to use for disposal. {\bf (4)} Once the user sees the LED color, they should throw the trash into the matching bin. {\bf (5)} Using proximity sensors at the rim of each bin, the system will detect when and where the trash item is disposed and send a confirmation message. Finally, {\bf (6)} if the trash item was disposed in the correct bin, the system will send a reward to the user.

\subsection{Mechanical Design}
\label{subsec:3d}

\begin{figure}[tbh]
\centering
\includegraphics[width=0.9\columnwidth]{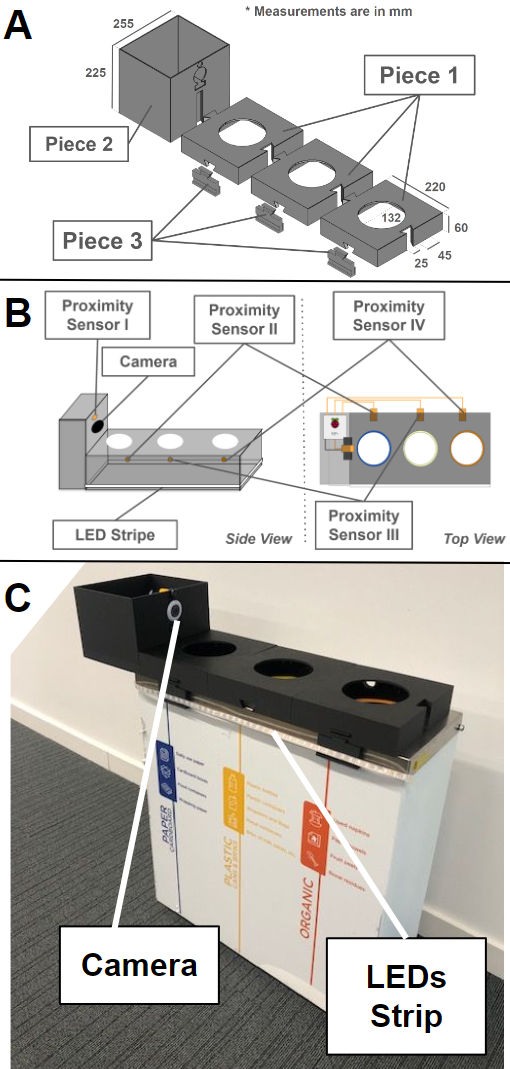}
\caption{Overview of iTrash design and components. A) 3D mechanical components, exploded-axonometric view. B) Hardware components and connections, side and top views. C) Installation on a real trashcan.} 
\label{fig:itrash}
\end{figure}

iTrash functions as a robotic system add-on to a standard trashcan (see Fig.~\ref{fig:itrash}). It can be adapted to different trashcans by making minor adjustments to the design of the 3D-printed mechanical parts (the other components remain the same) and is easy to install and remove, allowing for quick repairs. 

We designed the mechanical components to be produced using widely available desktop-size 3D printers, using PLA filament (we used a Bambu Lab X1-Carbon printer). To accommodate standard printing bed sizes, we divided the mechanical design into several pieces to be assembled (see Fig.~\ref{fig:itrash} A). {\bf Piece 1} fits above a bin opening, with a mounting position for a proximity sensor, as seen in Fig.~\ref{fig:itrash} B. In our setup, the base trashcan has 3 bins, so we use three instances of Piece 1. {\bf Piece 2} is a casing that holds the single-board computer (Raspberry Pi 4B), a camera, and an additional proximity sensor. Piece 2 positions the camera and proximity sensor above the other pieces, for a suitable object detection view. {\bf Piece 3} attaches the other components to the trashcan underneath and includes a mounting position for the LED strip (see Fig.~\ref{fig:itrash} C). All three of the piece types include dovetail joints to be secured together. 

\subsection{Hardware}
\label{subsec:hardware}
The hardware components in the iTrash system include four proximity sensors, an RGB LED strip, a camera, and a single-board computer. The {\bf proximity sensors} are used to detect when the user places an item near the trashcan (Proximity Sensor I, in Fig.~\ref{fig:itrash} B) and to detect into which bin the trash is disposed (Proximity Sensors II, III, and IV). We use a E18-B03P1 proximity sensors with a detection range of 5 to 30 cm. The {\bf RGB LED strip} is used to provide visual feedback to users, indicating that the system is ready, processing, or signaling the correct bin to place items. We use a WS2813 1-meter long strip with 60 LEDs. The {\bf camera} is used to capture images of waste items as they are detected by Proximity Sensor I, and to classify the items for proper sorting and recycling using computer vision. We use a Samzuy 1080P USB HD camera. For the {\bf single-board computer} to control the above components, we use a Raspberry Pi 4B. The proximity sensors and the LED strip are each connected to GPIO pins and the camera module is interfaced via a USB connection. To power the components, the setup employs both 5V power pins and two ground pins.

\subsection{Software}
\label{subsec:software}

\begin{figure}[tbh]
\centering
\includegraphics[width=0.605\columnwidth]{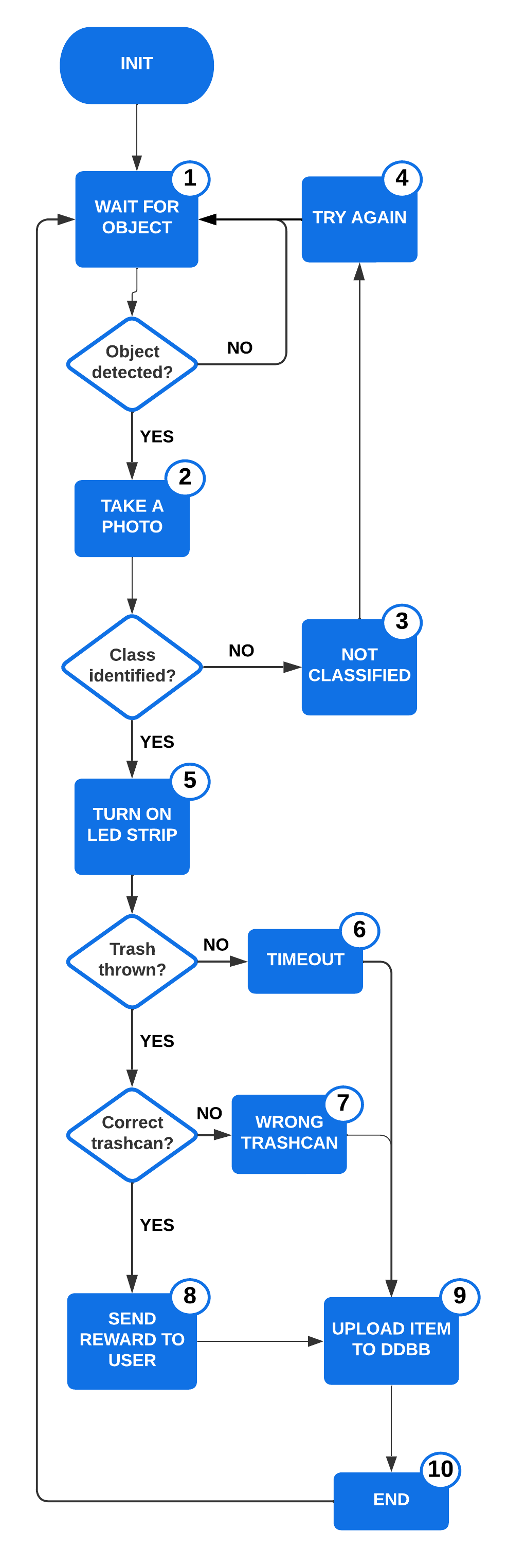}
\caption{State machine diagram for the iTrash software controller. The diagram shows the information flow of the system when a person interacts with iTrash.} 
\label{fig:controller}
\end{figure}

The state machine of the iTrash controller in shown in Fig.~\ref{fig:controller}. 
Initially, {\bf (1)} iTrash waits until the proximity sensor by the camera is triggered (Proximity Sensor I, in Fig.~\ref{fig:itrash} B), indicating that a user is holding something in front of the camera. When the proximity sensor is triggered, {\bf (2)} iTrash captures an image with the camera, then tries to classify any object in the image. {\bf (3)} If an object in the image is not successfully classified, for example because an item was moved too quickly or the image is unclear, {\bf (4)} the user will be asked to try again. If an object is successfully classified, {\bf (5)} the system will illuminate the RGB LED strip in the color matching the appropriate bin for the classified object, thus informing the user. Then, the user is supposed to dispose the item in the indicated bin, thus triggering the proximity sensor above the correct bin's opening (one of Proximity Sensors II, III, and IV, in Fig.~\ref{fig:itrash} B). If none of the proximity sensors of the bin openings are triggered within 10 seconds, {\bf (6)} the system will timeout and the process will need to be restarted. {\bf (7)} If the proximity sensor of a bin that was not indicated is triggered, the bin is considered incorrect and the user is not rewarded. If the proximity sensor of the indicated bin is triggered, {\bf (8)} the user receives a reward. In all three cases (timeout, incorrect bin, or correct bin with reward), the image taken by the camera {\bf (9)} is stored in a database for machine learning and quality assessment purposes. Finally, {\bf (10)} the process will end, and the system will return to its initial state. The controller\footnote{All the code used in this project can be found on its Github repository: \url{https://github.com/IERoboticsAILab/iTrash}} was implemented in Python, with MongoDB\footnote{\url{https://www.mongodb.com}} used for the database.

\subsection{Classification Model}
\label{subsec:classificationmodel}

iTrash uses a classification model to determine which bin to recommend to the user. Following a trial-and-error testing period with a variety of models and datasets, we selected gpt-4o-mini~\cite{gpt40mini} as the model to use. The input of the model is an image of the item the user wants to dispose. If the image is valid (i.e., with an item in the foreground), the model will decide between three possible classification outputs: blue (paper and cardboard), yellow (plastic) and brown (organic), otherwise, the model will require the user to display the item again.

For each item iTrash classifies, the following parameters are stored for analysis: 
\begin{itemize}
    \item \textit{image}: stored in base64 format, which optimizes space usage and allows easy decoding to PNG format whenever needed for review. 
    \item \textit{time}: date and time at which the item associated to the image was disposed, or at which the process timed out.
    \item \textit{{bin\_predicted}}: the color value of the bin predicted by the model for the item.
    \item \textit{{bin\_thrown}}: the color value of the bin where the user disposed the item.
\end{itemize}

The parameters \textit{bin\_predicted} and \textit{bin\_thrown} are used not only to assess the rate of users that follow the indicated instruction, but also the rate of iTrash's correct predictions. To assess the rate of correct predictions, we use a parameter \textit{{bin\_real}}, which is the color value of the real bin into which the article should have been classified according to manual quality review. Ideally, in an experiment, the three values will line up, indicating that iTrash classified the item correctly and the user followed the instructions.

\subsection{Blockchain}
\label{subsec:blockchain}
The reward system is implemented using blockchain technology. Users interacting with iTrash are expected to have already created an account linked to a wallet on the Ripple (XRP) network. All the transactions issued in this research occurred in the XRP Testnet\footnote{\url{https://xrpl.org/resources/dev-tools/xrp-faucets}}.

When iTrash is ready to reward a user, it displays a message on a connected LCD screen asking the user to present the QR code of their wallet on the XRP Testnet. If the user presents their QR code in front of iTrash's camera within a 10 second timer, the reward is sent to the user's wallet. If no QR code is presented, iTrash displays four NGO options on the connected LCD screen, so the user can select one to donate their reward. Each displayed NGO is accompanied by a QR code, which the user scans to select the option.

To implement this, we established five wallets in the Testnet of Ripple\footnote{\url{https://xrpl.org/es-es/resources/dev-tools/xrp-faucets}}: one for iTrash and four for the respective NGOs. We developed a Flask app hosted on Google App Engine\footnote{\url{https://cloud.google.com/appengine?hl=es_419}} and created four endpoints, one for each NGO. When a user scans a QR code and accesses the corresponding endpoint, the system transfers 0.01 XRPs to the selected NGO's wallet from the iTrash's wallet and a message stating 'Reward sent!' appears on the connected LCD screen.

\section{Experiment setup}
\label{sec:experiments}

\begin{figure}[tbh]
\centering
\includegraphics[width=\columnwidth]{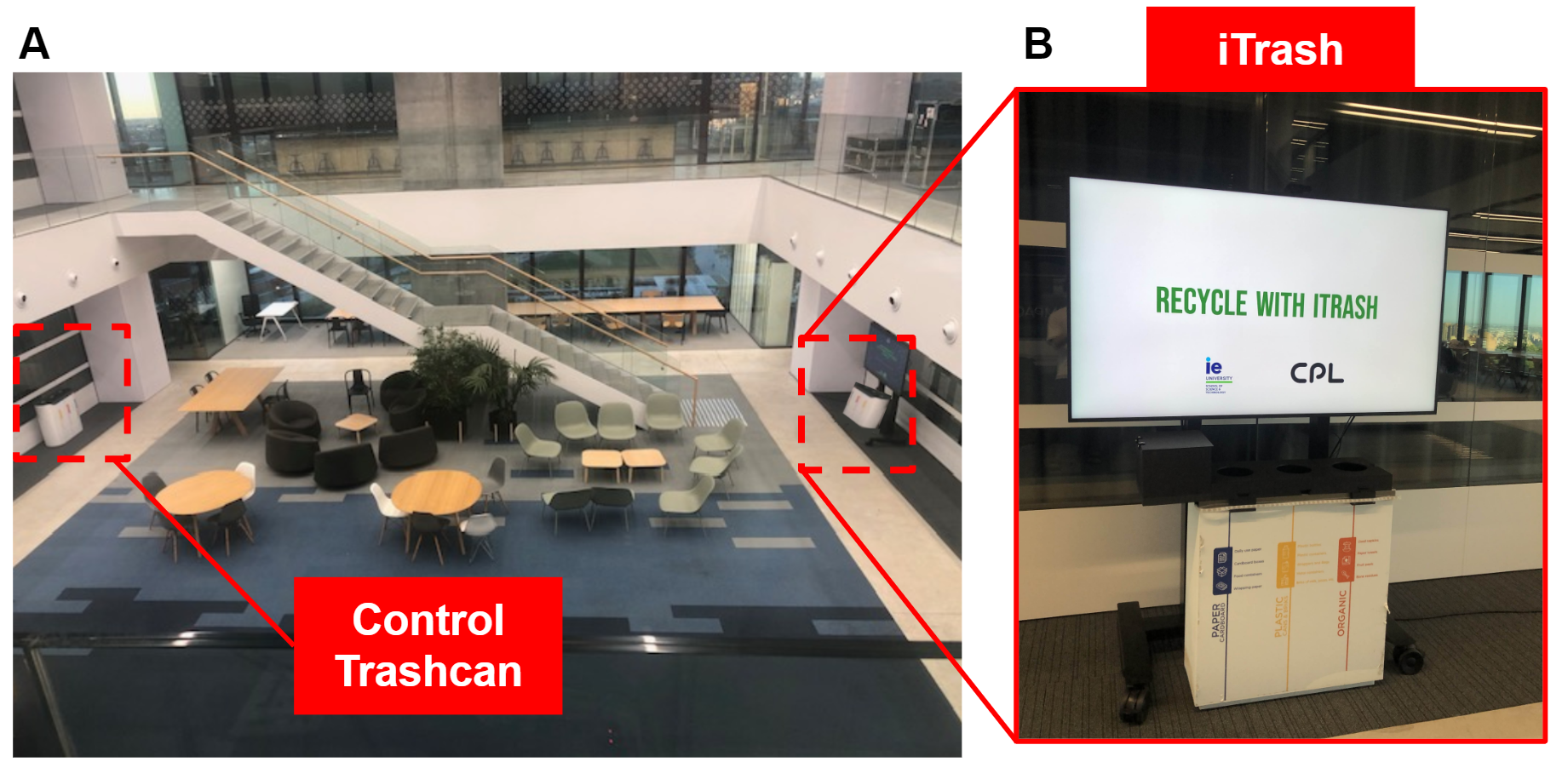}
\caption{Environmental setup for the proposed system. A) Panoramic view of the area where the experiments were performed, in the right-hand side an iTrash is located, opposite to it, the control trashcan can be seen. The separation between these two devices was 25 meters. B) \textit{iTrash} together with an additional display screen where the instructions to use the device were shown. } 
\label{fig:experiments}
\end{figure}

To demonstrate the feasibility of iTrash (see Fig.~\ref{fig:experiments}), we ran a 5-day experiment comparing the accuracy of iTrash against a normal (control) trashcan. The two trashcans (iTrash and control) were placed on opposite sides of a spacious (approximately 250 $m^{2}$) indoor common area in a university campus, where many potential users, including students and staff, passed by both systems every day. As depicted in Fig.~\ref{fig:experiments}, we connected iTrash to an LCD screen to display visual support for the user. When no object has recently been detected by iTrash, a 50-second video\footnote{\url{https://youtu.be/sdrd5JMhjsk}} explaining how the system works plays on a loop. 

Each of the experimental days (Monday to Friday) began at 8 AM, with both trashcans (iTrash and control) empty. The trashcans remained in place for 12 hours. Each experimental day ended at 8 PM, when the trashcans were emptied and the data collected.
For iTrash, we manually reviewed all of the images taken during that day and determined into which bin each item should have been disposed, according to the standards set by the local city council\footnote{\url{https://www.comunidad.madrid/servicios/consumo/reciclaje-donde-tirar-cada-producto}}. These manual determinations were recorded in the parameter \textit{bin\_real} for each item. For the control, we manually reviewed each item in each bin, recording the bin the item was found in as the \textit{bin\_thrown} and the bin into which the item should have been disposed as the \textit{bin\_real}. After 5 days, we had collected the predicted and real values for all items disposed in both trashcans.

\section{Results}
\label{sec:results}
\begin{figure}[h!]
\centering
\vspace{0.2cm}
\includegraphics[width=\columnwidth]{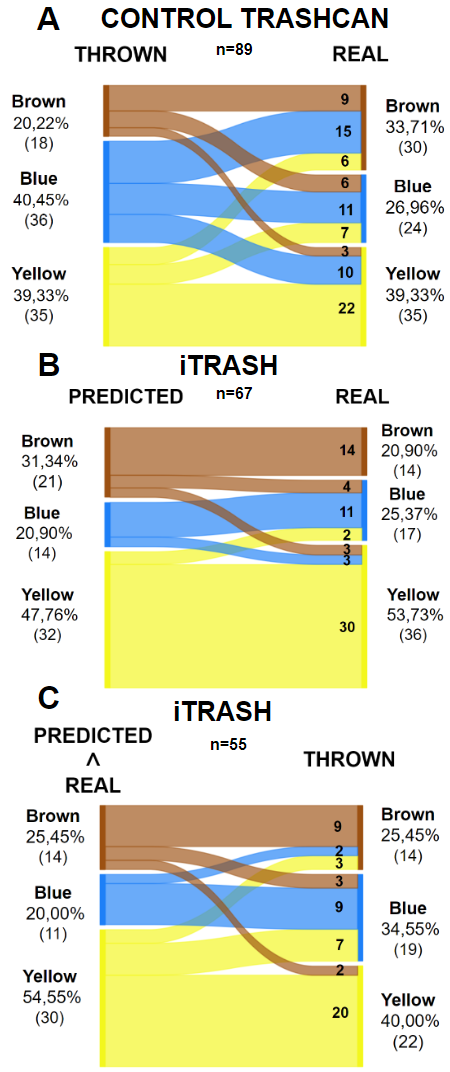}
\caption{Sankey diagrams of the results. A) Comparison of the thrown (\textit{bin\_thrown}) and real (\textit{bin\_real}) values in the control. B) Comparison of the predicted (\textit{bin\_predicted}) and real (\textit{bin\_real}) values in iTrash. C) Comparison of items having predicted values that match their real values (\textit{bin\_predicted} \(\land\) \textit{bin\_real}) and the thrown (\textit{bin\_thrown}) values in iTrash. The blue bin is for paper items, the yellow bin for plastic, and the brown bin for organic.}
\label{fig:sankeys}
\end{figure}

In the 5-day experiment, 89 total items were disposed in the control (36 in the blue bin, 18 in the brown bin, and 35 in the yellow bin) and 67 total items were disposed in iTrash (25 in the yellow bin, 24 in the blue bin, and 18 in the brown bin).

We report the results in terms of the \textit{bin\_predicted}, \textit{bin\_real} and \textit{bin\_thrown} values in Fig.~\ref{fig:sankeys}, using Sankey diagrams to visualize the flow from one category to another as well as some patterns in the error.

For the control, we compared the \textit{bin\_thrown} value with the \textit{bin\_real} value (see Fig.~\ref{fig:sankeys} A). We did not include the \textit{bin\_predicted} value for the control because it is equivalent to \textit{bin\_thrown}: the user disposes the item where the user predicts it should be disposed. In the control, of all the items that should have gone in the brown bin, 9 were correctly sorted, while 6 were placed in the blue bin and 3 in the yellow bin. For items that should have gone in the blue bin, 11 were correctly sorted, while 15 were placed in the brown bin and 10 in the yellow bin. For items that should have gone in the yellow bin, 22 were correctly sorted, while 6 were placed in the blue bin and 7 in the brown bin. 

For iTrash, we first compared the \textit{bin\_predicted} and \textit{bin\_real} values to evaluate the performance of the model (see Fig.~\ref{fig:sankeys} B). All items that should have gone in the brown bin were correctly predicted as brown. For items that should have gone in the blue bin, 11 were correctly predicted, while 4 were misclassified as brown and 2 as yellow. For items that should have gone in the yellow bin, 30 were correctly predicted, while 3 were misclassified as blue and 3 as brown.

Then, for iTrash cases in which the \textit{bin\_predicted} and \textit{bin\_real} values matched (i.e., iTrash predicted correctly), we compared the \textit{bin\_predicted} $\land$ \textit{bin\_real} and the \textit{bin\_thrown} values, to evaluate the rate of user cooperation with the robotic system (see Fig.~\ref{fig:sankeys} C). In total, there were 55 items that iTrash predicted correctly. Of the items correctly predicted as brown, 9 were correctly placed in the brown bin, while 3 were placed in the blue bin and 2 in the yellow bin. For the items correctly predicted as blue, 9 were correctly placed in the blue bin, while 2 were placed in the brown bin. Finally, for the items correctly predicted as yellow, 20 were correctly placed in the yellow bin, while 3 were placed in the brown bin and 7 in the blue bin.

\subsection{Accuracy comparison}
To calculate the accuracy of the iTrash classification model, we use:
\begin{equation}
        \textit{Accuracy} = \frac{\textit{Correct predictions}}{\textit{All predictions}},
    \label{eq:accuracy}
\end{equation}
where \textit{Correct predictions} refers to the number of items where \textit{bin\_predicted} and \textit{bin\_real} values matched. In the control, the users achieved a prediction (and disposing) \textit{Accuracy} of 47.19\%, while iTrash achieved a prediction \textit{Accuracy} of 82.09\%.  
\section{Discussion}
\label{sec:discussion}

\begin{figure*}[tbh]
\centering
\includegraphics[width=\textwidth]{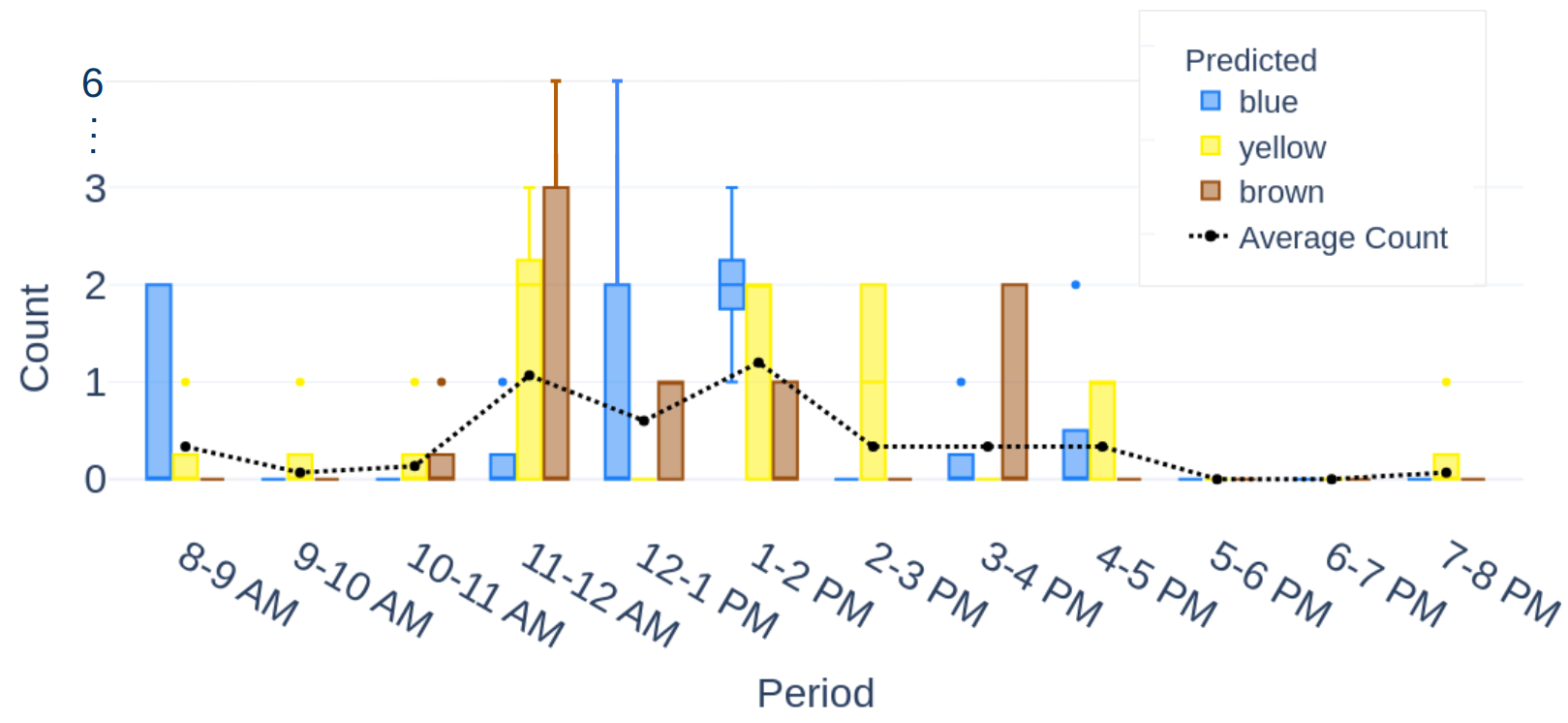}
\caption{Overview of the number of items disposed of per hour in each bin throughout the experiment. Each time slot is represented by three boxplots, one for each bin (blue, yellow, and brown). TThe boxes illustrate the interquartile range (IQR) of the number of items discarded in each bin. The whiskers cover values within 1.5 times the IQR of the lower and upper quartiles. Any points beyond this range are considered outliers, indicating instances where the number of discarded items significantly deviated from the typical pattern.The dotted line represents the average count of disposed items per time slot.} 
\label{fig:time_analysis}
\end{figure*}

In the analysis of the results, we have excluded items that were presented to the iTrash camera but then not disposed in any bin. A total of 79 items were presented to the iTrash camera and recorded in the database, and 12 of these were not disposed (15\% of the items). Manually reviewing the images showed that most of these items were phones, wallets, or earphones, thus, users were likely experimenting with how the system worked. We also did not include the reward data in the analysis of the results, because only 2 users chose one of the NGOs to receive the reward they earned, suggesting that there might have been some misunderstanding among the users about how the reward system worked, or that users were reluctant to spend more time interacting with the system. 

Note that, in Fig.~\ref{fig:sankeys}, an ideal system with 100\% accuracy would result in all straight, aligned bars for all colors. Fig.~\ref{fig:sankeys} and the \textit{Accuracy} values (Eq.~\ref{eq:accuracy}) of the results show that iTrash's predictions had a lower rate of error than the control. Specifically, the average prediction of iTrash outperformed the average user's choice in the control by 34.9\%. This suggests that using iTrash provides a strong improvement in recycling accuracy compared to users relying on their own judgment. 

In the control (see Fig.~\ref{fig:sankeys} A), the bin with the most incorrect items thrown into it was the blue bin: less than one-third of the items placed in the blue bin were correctly sorted. Conversely, the bin with the most correct items thrown into was the yellow bin, with an accuracy rate of approximately 63\%. This suggests that users were more aware of the regulations about which items should be disposed of in the yellow bin, or at least found the regulations less ambiguous to follow. Indeed, it should be straightforward to identify which items are plastic, with the exception of coffee cups, which are composed of both paper and plastic and therefore belong in the yellow bin according to the regulations.

In iTrash (see Fig.~\ref{fig:sankeys} B), the bin with the most incorrect predictions was the brown bin (that is, blue and yellow items being incorrectly predicted as belonging to the brown bin). However, no brown items were incorrectly predicted as belonging to the blue or yellow bins. This suggests that the iTrash classification model currently has a slight bias to classify items as brown. In addition to this slight bias towards brown, the iTrash classification model occasionally confuses blue and yellow items (5 items incorrectly swapped between yellow and blue, out of 46 items predicted to be yellow or blue).

When iTrash made correct predictions (see Figure.~\ref{fig:sankeys} C), users usually followed iTrash's instructions (almost 70\% of the time), but sometimes override them. When iTrash's predictions were correct for the blue bin, users followed the instruction to throw an item in the blue bin more than 80\% of the time. However, when iTrash's predictions were correct for the yellow bin, users incorrectly overrode iTrash's instructions one-third of the time. It is worth noting that when iTrash made incorrect predictions, users usually followed the wrong instructions, rather than correctly overriding them.

A potential additional use case of the iTrash system is to track bin usage throughout the day in busy common areas. Fig.~\ref{fig:time_analysis} reports the number of items disposed of by color, grouped by time periods over the course of the five experimental days. The boxplots in the figure provide a detailed statistical representation of waste disposal events for each bin color in each time slot. Each box represents the interquartile range (IQR) of the number of items discarded in each bin during each time slot, over the course of the 5-day experiment, capturing the middle 50\% of the data (from the $25^{th}$ percentile to the $75^{th}$ percentile). The whiskers represent data dispersion, covering values within 1.5 times the IQR from the lower and upper quartiles. Points beyond this range are considered outliers, representing instances where the number of discarded items significantly deviated from the typical pattern. Notably, the blue boxplot for the 1-2 PM period shows a lower whisker that does not reach zero, suggesting that at least one item was always disposed during this period across all experimental days.

The results indicate a peak in waste disposal during midday (11:00 to 14:00), where the median and upper quartiles show a marked increase, almost doubling compared to the rest of the day. This pattern is consistent with typical cafeteria usage, where meal times generate much more waste due to increased food consumption and packaging disposal. 
\section{Future Work}
For future work, we would like to focus on analyzing the temporal data produced by iTrash (see Fig.~\ref{fig:time_analysis}). For instance, conducting a temporal analysis of household garbage disposal could provide valuable information about personal habits. Monitoring the frequency and timing of waste disposal can help identify peak periods, such as after meals or during routine cleanups. This data could be useful for optimizing municipal waste management systems, such as scheduling garbage collections, for organizing waste management in high-traffic areas with many small bins (office buildings, universities), or even for integrating smart technology into household bins, not only to improve waste classification accuracy but to potentially provide recommendations for more environmentally-friendly consumption patterns.

Another possible line of experimentation is improving the reward system by incorporating additional sensors (e.g., a weight sensor), allowing the reward amount to be proportional to the weight of the disposed item. Also, regarding the classification model, we could build a dataset using images stored by iTrash, continuously improving its classification accuracy.

Regarding the blockchain-based reward system, due to the low participation rate that we got in the experiments, we should consider a different setup where users directly receive the reward and are provided with options to spend it on products or discounts. Another factor that could explain the low scanning rates is the reluctance of some users to spend extra time interacting with the system. Therefore, we should explore alternative methods of delivering the reward. The primary challenge is identifying the user, and one potential solution could be using NFC technology. The idea would be for users to have a blockchain wallet, and once they recycle correctly, they could use their phone or an NFC tag to scan, allowing the reward to be sent directly to their wallet. This approach will save users time by eliminating the need to open their camera and scan a QR code or open their wallet app and show a QR code to the camera. Increasing the accessibility and ease of receiving rewards might increase the rate of users that follow iTrash's instructions instead of overriding them, especially if they frequent the space and become repeat users.

\section{Conclusions}
\label{sec:conclusions}
This article introduces iTrash, a smart trash can system that aims to improve recycling efficiency through computer vision and blockchain-based incentives. In a 5-day experiment, iTrash achieved 34.9\% more accuracy than traditional trash cans, demonstrating its effectiveness in reducing human error through automatic waste sorting. This can in part be attributed to iTrash's rate of prediction accuracy being much higher than the average human user's, but might also be partly attributed a high level of trust in the technology or a high rate of users being comfortable relying on robotics systems to make this type of decision, as users largely followed iTrash's recommendations even with a very low rate of users choosing to collect the reward that was offered. Although the blockchain reward mechanism showed potential, low participation suggests the need for more seamless reward delivery, such as NFC-based systems. In conclusion, iTrash effectively improves recycling accuracy in small environments such as offices and universities, offering a scalable solution for smarter and more sustainable waste management.

\section{Acknowledgments}
Project supported by a 2024 Leonardo Grant for Scientific Research and Cultural Creation from the BBVA Foundation. The BBVA Foundation accepts no responsibility for the opinions, statements and contents included in the project and/or the results thereof, which are entirely the responsibility of the authors.

\bibliographystyle{IEEEtran}
\bibliography{references}

\end{document}